\documentclass[lettersize,journal]{IEEEtran}
\usepackage{amsmath,amsfonts}
\usepackage{array}
\usepackage[caption=false,font=normalsize,labelfont=sf,textfont=sf]{subfig}
\usepackage{textcomp}
\usepackage{stfloats}
\usepackage{url}
\usepackage{verbatim}
\usepackage{graphicx}
\usepackage{cite}
\usepackage{comment}
\usepackage{tabularx}
\usepackage{caption}
\usepackage{hyperref}
\usepackage{xcolor}
\newcommand{\etal}{\textit{et al}.}
\newcommand{\ie}{\textit{i}.\textit{e}. }

\usepackage{algorithm,algcompatible,amsmath}
\usepackage{amssymb}
\algnewcommand\INPUT{\item[\textbf{Input:}]}%
\algnewcommand\OUTPUT{\item[\textbf{Output:}]}%
\usepackage{booktabs} 
\usepackage{pbox}
\usepackage{adjustbox}
\usepackage{colortbl}
\definecolor{Gray}{gray}{0.85}
\newcolumntype{a}{>{\columncolor{Gray}}c}
\newcolumntype{b}{>{\columncolor{white}}c}
\begin{document}

\title{Dual Stage Stylization Modulation for Domain Generalized Semantic Segmentation}

\author{Gabriel Tjio,
        Ping Liu*,  Chee Keong Kwoh
        and Joey Tianyi Zhou
\thanks{This paper was produced by CFAR, Agency of Science, Technology and Research.}
\thanks{Manuscript received August 03, 2023; 
Gabriel Tjio, Ping Liu, and Joey Tianyi Zhou are with the Center for Frontier AI  Research, A*STAR.  (email:liu\_ping@cfar.a-star.edu.sg, gabriel-tjio@cfar.a-star.edu.sg,joey\_zhou@cfar.a-star.edu.sg)  \\ 
Kwoh Chee Keong is with the Nanyang Technological University (email:asckkwoh@ntu.edu.sg) \\
 * Ping Liu is the corresponding author. \\
 \ }}

\markboth{Journal of \LaTeX\ Class Files,~Vol.~14, No.~8, August~2021}%
{Shell \MakeLowercase{\textit{et al.}}: A Sample Article Using IEEEtran.cls for IEEE Journals}

\IEEEpubid{0000--0000/00\$00.00~\copyright~2021 IEEE}
\maketitle

\begin{abstract}
Obtaining sufficient labeled data for training deep models is often challenging in real-life applications. To address this issue, we propose a novel solution for single-source domain generalized semantic segmentation. 
Recent approaches have explored data diversity enhancement using hallucination techniques. 
However, excessive hallucination can degrade performance, particularly for imbalanced datasets. 
As shown in our experiments, minority classes are more susceptible to performance reduction due to hallucination compared to majority classes.
 To tackle this challenge, we introduce a \textbf{d}ual-stage \textbf{F}eature \textbf{T}ransform ($dFT$) layer within the \textbf{A}dversarial \textbf{S}emantic \textbf{H}allucination+ (ASH{+}) framework. 
The ASH{+} framework performs a {dual-stage manipulation of hallucination strength}. 
By leveraging semantic information {for each pixel}, our approach adaptively adjusts the pixel-wise hallucination strength, thus providing fine-grained control over hallucination.
We validate the effectiveness of our proposed method through comprehensive experiments on publicly available semantic segmentation benchmark datasets {(Cityscapes and SYNTHIA)}. Quantitative and qualitative comparisons demonstrate that our approach is competitive with state-of-the-art methods for the Cityscapes dataset and surpasses existing solutions for the SYNTHIA dataset. Code for our framework will be made readily available to the research community.
\end{abstract}


\section{Introduction}
\IEEEPARstart{D}{}eep learning has significantly advanced the state-of-the-art performance for various computer vision tasks, including image classification \cite{imagenet_classification}, semantic segmentation\cite{dec_attn_zhang2019}, image retrieval \cite{dynamic_contrast_rao2023} and image editing \cite{spg_2022}. However, achieving such performance gains often requires a large amount of labeled training data. Meeting this requirement is challenging for many real-world applications \cite{YAKIMOVICH2021100383, wang2020weaklysupWSL}, particularly those that {require} expert knowledge for data labeling.

Moreover, manual data annotation can be labor-intensive, especially for semantic segmentation tasks, which typically require pixel-level annotations. While using synthetic data for training {reduces} annotation costs, it often leads to a domain shift \cite{dgsurvey} between the synthetic data (source domain) and real-world {data} (target domain). This domain shift refers to the distributional differences between the source and target domain data, which significantly impacts the performance on the target domain data.

\begin{figure*}
\centering
\includegraphics[width=0.9\textwidth,keepaspectratio]{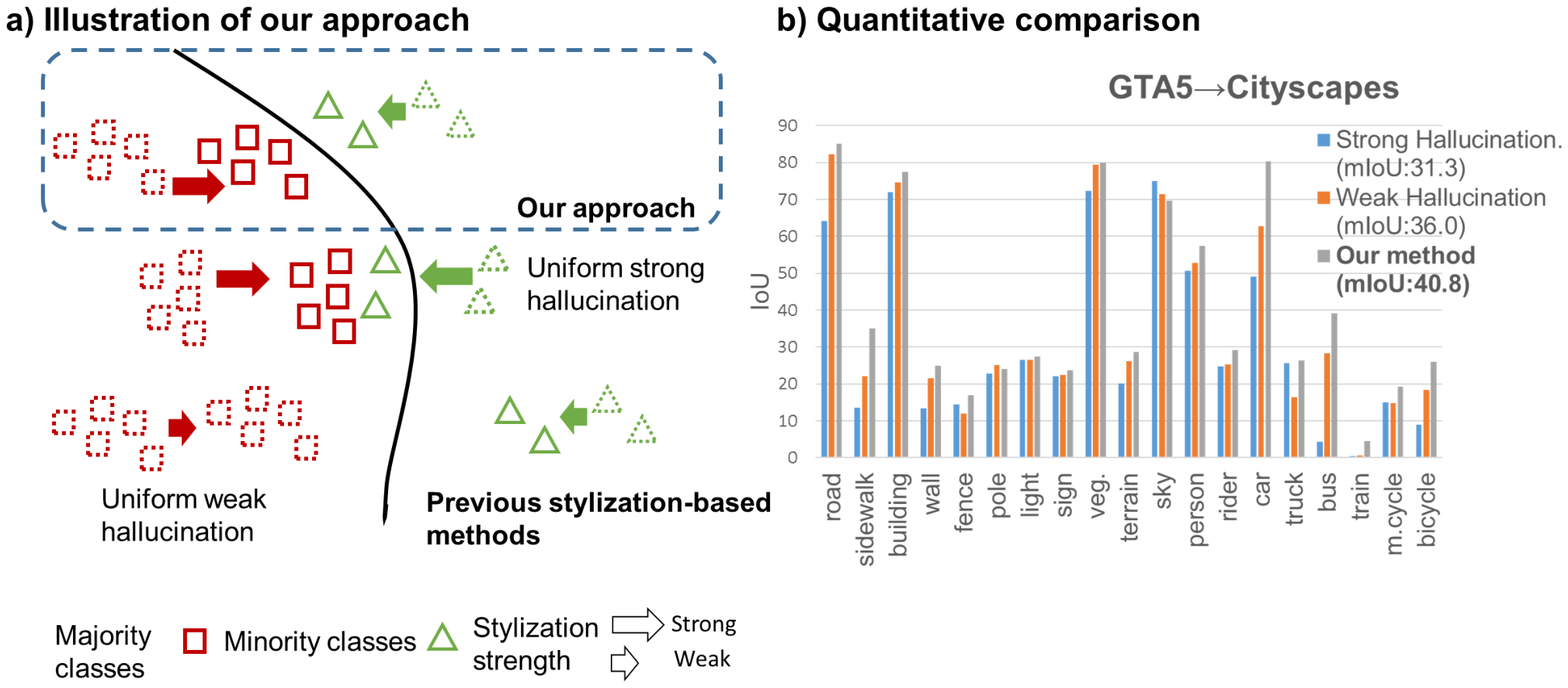}
\captionsetup{width=0.9\textwidth}
\caption{{Illustration of the motivation of our proposed solution, \textbf{A}dversarial \textbf{S}emantic \textbf{H}allucination+ (ASH{+}).
{Strong/weak hallucination refers to the stylization strength of the source domain images.}
Uniform hallucination of the source domain data risks worsening performance, {particularly for minority classes}, if class-wise properties are ignored.
{This can happen when the model overfits to the strongly augmented {instances from the} minority classes, ultimately shifting the decision boundary away from the correct decision boundary.}
In contrast, ASH{+} overcomes these limitations by conditioning the stylization process with semantic information. 
By employing semantically-aware stylization, we account for the varying classification difficulty across different classes. 
This class-aware stylization approach improves the overall model performance and demonstrates the importance of considering {the pixel-wise semantic differences.}
}}
\label{figure:ash_vs_stylization}
\end{figure*}

While leveraging unlabelled target domain data \cite{DBLP:conf/nips/LuoLGY020,DBLP:series/acvpr/GaninUAGLLML17,Luo_2019_CVPR,NEURIPS2018_ab88b157,9616392,Gao2022adv_uda_actionrecog} can effectively minimize the performance drop caused by the domain shift, it is worth noting that target domain data may not always be available during training due to data privacy concerns or other limitations. As a result, domain generalization methods aim to address the challenge of domain shift when access to target domain data is restricted.

Domain generalization approaches can be broadly categorized into two groups: single-source domain generalization \cite{Yue_2019_ICCV,Li_2020_CVPR,uzair2017_blindDA,li_genknow2022,jin_stylenorm} and multiple-source domain generalization \cite{zhou2021mixstyle,kd_dg_2023,liu_enc_rs_2023}. In the former approach, a model is trained using a single labeled source domain dataset, while the latter approach utilizes multiple labeled datasets to enhance generalization capabilities. 
In our work, we focus specifically on single-source domain generalization. 
This choice is motivated by the practical challenges associated with collecting multiple source domain data, which can be difficult or even infeasible in certain scenarios.
\IEEEpubidadjcol

Recently, there has been increasing attention on hallucination-based approaches\cite{Li_2020_CVPR,DBLP:conf/nips/LuoLGY020,Yue_2019_ICCV}, which aim to enhance the performance of single-source domain generalization by increasing the  training data diversity. 
These approaches \cite{Li_2020_CVPR,DBLP:conf/nips/LuoLGY020,Yue_2019_ICCV} generate data through various data transformations. These transformations alter color and texture information while preserving shape and structure information within the source domain images.
The underlying idea behind these approaches is that randomly varying the domain-variant features while preserving the domain-invariant features {in the training data} reduces the spurious correlations between the domain-variant features and the ground truth labels. 
These spurious correlations may not generalize well to the target domain and may potentially reduce performance. 

While hallucination-based approaches aim to increase training data diversity, they have certain limitations that need to be considered. 
Some prior works\cite{Yue_2019_ICCV,Kundu_2021_ICCV} uniformly stylize the source domain images without considering the variations between pixels. 
This may lead to challenges in classifying certain pixels (Figure \ref{figure:ash_vs_stylization}a), particularly those belonging to minority classes.
Uniform data augmentation disproportionately affects minority classes compared to the majority classes because the model may overfit to the augmented minority class {instances}, resulting in a gap between the learned decision boundary and the correct decision boundary. 
Furthermore, in Figure \ref{figure:ash_vs_stylization}b), the results show that uniformly stylizing the source domain images from the GTA5 dataset \cite{Richter_2016_ECCV} reduces the performance depending on the hallucination strength.
Performance drops considerably with uniform strong hallucination (mIoU: 31.3). 
This shows that a large hallucination strength increases the task difficulty for the minority classes (Figure \ref{figure:ash_vs_stylization}b).
The performance for uniformly  weak hallucination strength (mIoU: 36.0) is similar to that of training with source-only data (mIoU: 36.6, Table \ref{tab:gta-cityscapes-paper-proof}) 
These observations motivate us to propose an adaptive hallucination approach that considers the characteristics of different classes, {with the aim of addressing} the limitations of uniform stylization. 

To address the aforementioned challenges in single-source domain generalization, we propose a novel approach termed \textbf{A}dversarial \textbf{S}emantic \textbf{H}allucination+ (ASH{+}). 
This method involves a dual-stage manipulation of hallucination strength that builds upon our previous work ASH \cite{Tjio_2022_WACV} and also presents substantial improvements.
In our earlier work, we employed a global predefined hyperparameter to uniformly balance the stylized source domain features with the original source domain features during the hallucination process. 
However, as discussed above, this uniform stylization may not be optimal and could degrade performance.
To overcome this limitation, we introduce the concept of a style-content balancing weight denoted as $\boldsymbol{\alpha}$. 
This weight modulates the stylization process based on the semantic information present in the images. 
By selectively controlling the {hallucination strength} for different classes
of the images, we can introduce significant stylistic variations depending on the input image.
The introduction of the style-content balancing weight represents a significant advancement over our previous work\cite{Tjio_2022_WACV} because $\boldsymbol{\alpha}$  enables a more fine-grained and adaptive manipulation of hallucination strength based on the semantic characteristics of the images.

Specifically, we propose an integrated dual-stage manipulation of hallucination strength framework that combines our previous work\cite{Tjio_2022_WACV} with the newly introduced style-content balancing weight $\boldsymbol{\alpha}$.
In the first stage of manipulation of hallucination strength, we increase the stylistic diversity via learned affine transformations of the style features. 
This exposes the model to a more diverse range of training data, which improves its ability to generalize to unseen domains.

In the second stage of manipulation of hallucination strength, we employ the style-content balancing weight $\boldsymbol{\alpha}$ to control the stylization extent in the generated images. 
This weight balances the proportions of the original source domain features and the stylized source domain features, effectively controlling the stylization strength at a pixel level. 
Notably, the style-content balancing weight $\boldsymbol{\alpha}$ is updated adversarially, allowing our approach to generate increasingly challenging examples during training. 
This dynamic manipulation of hallucination strength based on semantic information contributes to the generation of more diverse and informative training data, which ultimately benefits the model's generalization performance.
To demonstrate the {effectiveness of our latest update}, we conduct extensive experiments and provide detailed comparisons.
The results from these experiments demonstrate the {competitiveness} of ASH{+}.

Our main contributions are summarized below:
\begin{itemize}
    \item We propose a novel adversarial framework, named \textbf{A}dversarial \textbf{S}emantic \textbf{H}allucination+ (ASH{+}) for single-source domain generalization. 
    {We leverage semantic information to adaptively stylize the source domain images via dual-stage manipulation of hallucination strength. 
    The first stage allows us to increase the stylistic diversity of the stylized source domain images based on the semantic information of the source domain images while the second stage modulates stylization strength to account for class-wise differences across the images.} 
    \item We conduct extensive experiments to demonstrate the effectiveness of our approach. We evaluate our approach on the single-source domain generalization tasks with several benchmark datasets \ie GTA5 \cite{Richter_2016_ECCV} \slash Synthia\cite{RosCVPR16}\textrightarrow Cityscapes \cite{Cordts2016Cityscapes} and the SYNTHIA\cite{RosCVPR16} dataset. 
    ASH{+} demonstrates state-of-the-art performance for single-source domain generalization semantic segmentation. In particular, ASH{+} outperforms the state-of-the-art work\cite{qiaoCVPR20learning,Li_2021_CVPR} by as much as 38.48\% and 11.18\% respectively on the SYNTHIA\cite{RosCVPR16} dataset.

\end{itemize}

\section{Related Work}
\subsection{Domain Generalization}
Previous approaches in the field of single-source domain generalization can be categorized into two main categories: meta-learning methods \cite{Li2018learningGen, qiaoCVPR20learning} and data augmentation techniques \cite{uzair2017_blindDA, Zhang2020DST, Volpi2018NIPSAdvAug, shankar2018generalizing, zhou2021mixstyle, Yue_2019_ICCV, Xu_2021_CVPR}. 
These methods have been widely explored to enhance {model generalizability} when dealing with domain shift and limited labeled data. 
By leveraging meta-learning or data augmentation strategies, these prior works aim to improve the performance and robustness of single-source domain generalization models.

Meta-learning methods \cite{Li2018learningGen, qiaoCVPR20learning} enhance model generalizability by training on multiple tasks or domains to improve performance on unseen data. 
Li \etal \cite{Li2018learningGen} divide the source domain training data into meta-train and meta-test sets to simulate the domain shift between the source domain and the unseen target domain. 
Qiao \etal \cite{qiaoCVPR20learning} extend this approach by using adversarially augmented source domain data as meta-test data {to better approximate the} unseen target domain data. 
These meta-learning techniques seek to improve the model's ability to adapt to novel domains and achieve better generalization performance across different domains.

Data augmentation approaches \cite{uzair2017_blindDA, Zhang2020DST, Volpi2018NIPSAdvAug, shankar2018generalizing, zhou2021mixstyle, Yue_2019_ICCV,Xu_2021_CVPR} focus on diversifying the source domain data by varying domain-variant features. 
For example, Xu \etal \cite{Xu_2021_CVPR} specifically restrict data augmentation to domain-variant features by selectively transforming the low spatial frequency component in the images. 
These techniques aim to increase the diversity of the source domain data and alleviate overfitting to domain-specific features, ultimately improving the generalization performance of the model on unseen target domain data.

Unlike existing approaches, our method accounts for the inherent variations in task difficulty across different classes by dynamically varying the augmentation strength based on pixel-wise semantic information. 
This fine-grained hallucination strength modulation allows us to adaptively tailor the data augmentation to match the current model's capabilities and the specific characteristics of each class. 
By considering the semantic information among different classes, our method achieves a more targeted and effective data augmentation strategy, {which} further {enhances} the model's generalizability.

\subsection{Data Augmentation Via Style Transfer}
Data augmentation is commonly employed in machine learning to {increase} the {amount of} training data via label-preserving transformations such as color jittering \cite{redmon_yolo2016}, {randomized} image scaling \cite{simonyan2015deep}, and shifting \cite{redmon_yolo2016}. 
However, its effectiveness may vary depending on the specific context, as noted by Zeng \etal \cite{zeng2022reg_augdata}, and additional regularization measures are often required to ensure {generalizability} across diverse datasets.

In the context of domain adaptation, style transfer techniques have been explored to mitigate overfitting to domain-specific features by transferring styles from the target domain to the source domain \cite{pmlr-v80-hoffman18a, DBLP:conf/nips/LuoLGY020}. 
However, these methods rely on access to the target domain data and are not directly applicable to the domain generalization setting.

To tackle this challenge, several approaches have been proposed to enhance the diversity of the source domain data. These methods either randomize the existing source domain style information\cite{Zakharov_2019_ICCV,8202133} or transfer styles from auxiliary datasets \cite{Yue_2019_ICCV}. 
These techniques augment the source domain data by introducing style variations.

In contrast to these previous works, our approach adopts a generative-based method to perturb the styles and augment the data diversity while simultaneously adapting stylization strength. This approach enables us to generate synthetic samples with diverse styles, thereby enhancing the overall robustness and generalizability of the learned models.
\begin{figure*}[t]
\centering
\includegraphics[width=1.0\textwidth,keepaspectratio]{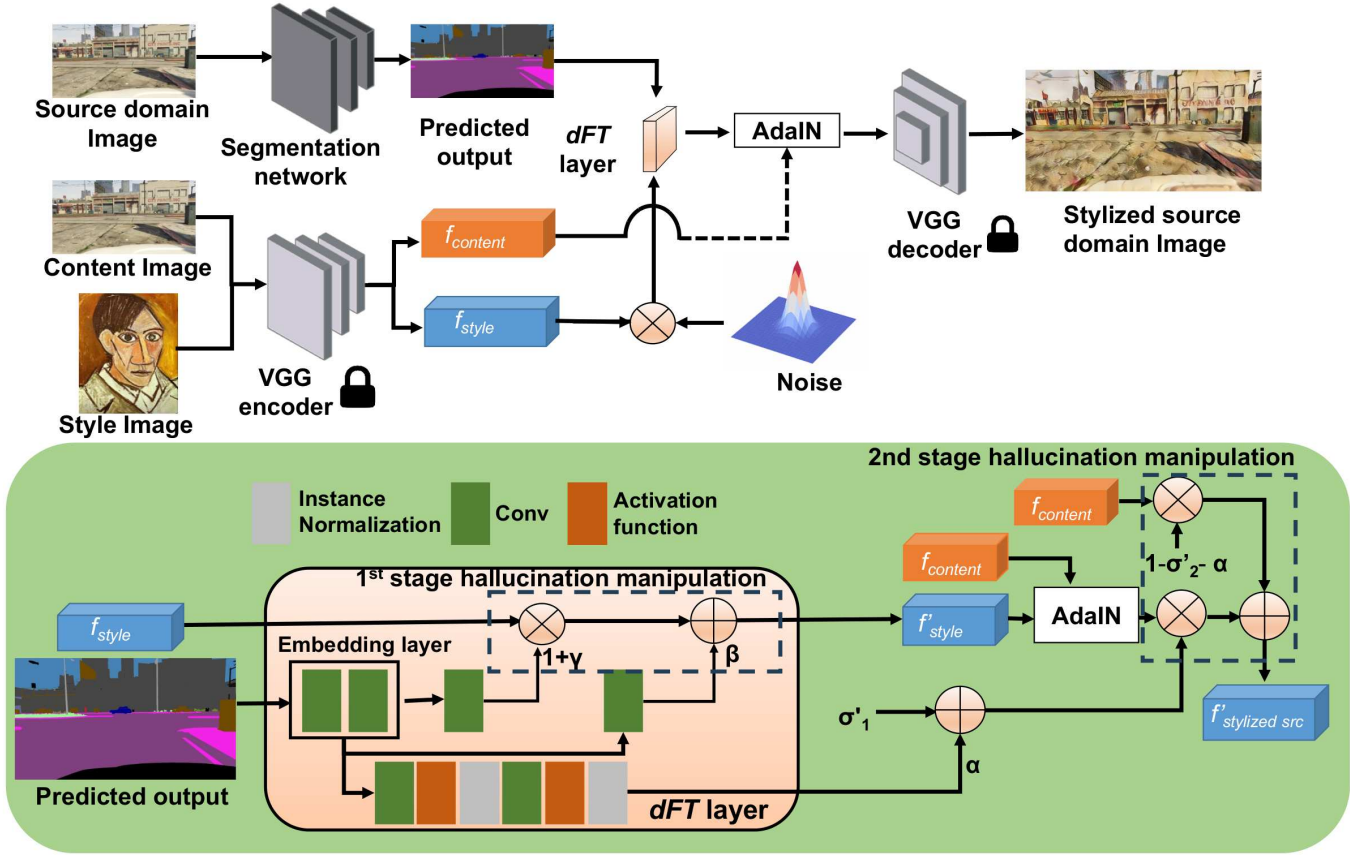}
\captionsetup{width=1.0\textwidth}
\caption{Description of our proposed workflow  \textbf{A}dversarial \textbf{S}emantic \textbf{H}allucination+ (ASH{+}). 
We use the predicted output from the source domain image, together with the style features from a randomly selected style image, as input to \textit{dFT} (\textbf{d}ual-stage \textbf{F}eature \textbf{T}ransform) layer. The \textit{dFT} layer then generates the element-wise scale $\boldsymbol{\gamma}$, shift $\boldsymbol{\beta}$ and style-content balancing weight $\boldsymbol{\alpha}$ variables. $\boldsymbol{\gamma}$ and $\boldsymbol{\beta}$ both perturb the input style features to introduce additional variation (the \textit{first} stage manipulation), while $\boldsymbol{\alpha}$ determines the element-wise proportion of the perturbed style features and content features in the reconstructed image (the \textit{second} stage manipulation). 
Different from earlier work \cite{Tjio_2022_WACV} which uses a global hyperparameter, the element-wise rebalancing variable $\boldsymbol{\alpha}$ allows for more fine-grained control over stylization. {$\boldsymbol{\sigma}^{'}_{1} , \boldsymbol{\sigma}^{'}_{2} $ are derived from the predefined hyperparameters $\sigma_{1},\sigma_{2}$ by scalar multiplication with a unit tensor. } }
\label{figure:workflow_fig}
\end{figure*}

\subsection{Style Transfer via Generative Models}

Generative models, particularly Generative Adversarial Networks (GANs) \cite{NIPS2014_5ca3e9b1}, have demonstrated their ability to generate diverse data that closely resembles the underlying training data distribution. 
Recent research\cite{wang2018sftgan,park2019SPADE,zhang2023semanticdehaze} has shown that incorporating semantic information as a prior can significantly improve the quality of synthesized images.

Wang et al. \cite{wang2018sftgan} introduced semantic information in the feature space during super-resolution, resulting in superior output image quality. 
By using the segmentation output from the low-resolution image as a prior, they controlled the transformation of low-resolution image features and preserved semantic features, leading to improved image quality. 
Similarly, Park et al. \cite{park2019SPADE} conditioned the generated image output with semantic information, which constrained stylistic variations within the semantic boundaries. 
This approach not only increased the realism of the generated images but also provided control over the layout within the generated images.

The effectiveness of using semantic information to enhance image quality is further supported by the recent work of Zhang et al. \cite{zhang2023semanticdehaze} in the context of image dehazing. 
However, while previous works \cite{wang2018sftgan, park2019SPADE, zhang2023semanticdehaze} utilized semantic information to condition image generation and enhancement, our approach leverages semantic information to generate challenging training data while preserving semantic details. 
\section{Method}
\begin{algorithm*}
  \caption{Adversarial approach for domain generalization}
 \begin{algorithmic}[1]
  \INPUT Source domain data $\boldsymbol{X_{s}}$, Source domain label $\boldsymbol{Y_{s}}$, Style image $X_{style}$, Segmentation network \textit {G}, Pretrained Encoder \textit {Enc}, Pretrained Decoder \textit {Dec}, dual Feature Transform \textit{dFT} layer , Adaptive Instance Normalization (AdaIN), Number of iterations $Iter_{num}$
  \OUTPUT Optimized segmentation network for domain generalization
  \FOR{$0$, ..., $Iter_{num}$}
   \STATE Obtain $\boldsymbol{f_{src}}$ via $Enc(\boldsymbol{X_{s}})$.
   \STATE Obtain $\boldsymbol{f_{style}}$ via $Enc(\boldsymbol{X_{style}})$. 
   \STATE Obtain $\boldsymbol{\alpha}$, $\boldsymbol{\gamma}$ and  $\boldsymbol{\beta}$ as shown in \textbf{Eq. \ref{eqn:classwise_scaleshift}}
   \STATE Perturb $\boldsymbol{f_{style}}$ with  $\boldsymbol{\gamma}$ and $\boldsymbol{\beta}$ as shown in \textbf{Eq. \ref{eqn:perturbed style features}} 
   \STATE Merge source features $\boldsymbol{f_{src}}$ and perturbed style features $\boldsymbol{f_{style}^{'}}$ with AdaIN.
   \STATE Apply rebalancing coefficient $\boldsymbol{\alpha}$ to balance the proportion of  $\boldsymbol{f_{src}}$ with the merged source-style features from step (6). 
   \STATE Generate stylized source image $\boldsymbol{X_{stylized}}$ by decoding the output from step (7) with $Dec$.  
   \STATE Train $dFT$ by minimizing $L_{ASH}(G,\boldsymbol{f_{src}},\boldsymbol{f_{style}^{'}})$.
   \STATE Train $G$ by minimizing $ \mathcal{L}_{seg}(G,\boldsymbol{X_{s}},\boldsymbol{Y_{s}})$ + $\mathcal{L}_{cont}(G,\boldsymbol{X_{stylized}},\boldsymbol{X_{s})}$. 
  \ENDFOR
 \end{algorithmic}
 \label{alg:workflow_alg}
\end{algorithm*}
In this section, we present the technical details of the Adversarial Semantic Hallucination+ framework (ASH{+}), which is designed to generate a diverse set of training data for domain generalizable models. 
The structure of this section is as follows: In Section \ref{subsection:prelim}, we provide a brief introduction to Adaptive Instance Normalization (AdaIN) \cite{huang2017adain} to establish the foundational concepts; In Section \ref{subsection:ASHplus_methods}, we elaborate on the \textit{dFT}(\textbf{d}ual-stage \textbf{F}eature \textbf{T}ransform) module architecture and discuss the objective functions used to guide the hallucination process; In Section \ref{subsection:trainingdetails} we focus on the training details of the \textit{dFT} layer, including the hyperparameters and optimization strategies employed.
\subsection{Preliminary Background}
\label{subsection:prelim}
Adaptive Instance Normalization (AdaIN) \cite{huang2017adain} is a technique used to {generate a new image containing the semantic information from a content image and the style information from a style image.} 
The process involves extracting features from both the content and style images using a pre-trained VGG19 \cite{Simonyan15} encoder. 

The AdaIN operation re-normalizes the channel-wise mean $\mu(.)$ and variance $\sigma(.)$ of the content features (denoted as $\boldsymbol{f_{src}}$) to match those of the style features $\boldsymbol{f_{style}}$. 
This allows the output features to align with the desired style while preserving the underlying content. 

Mathematically, the AdaIN operation can be expressed as follows:
\small
\begin{equation}
\text{AdaIN}(\boldsymbol{f_{src}}, \boldsymbol{f_{style}}) = \sigma(\boldsymbol{f_{style}}) \left( \frac{\boldsymbol{f_{src}} - \mu(\boldsymbol{f_{src}})}{\sigma(\boldsymbol{f_{src}})} \right) + \mu(\boldsymbol{f_{style}}), 
\label{eqn:adain}
\end{equation}
\normalsize

\subsection{Adversarial Semantic Hallucination+}
\label{subsection:ASHplus_methods}

Our proposed semantic hallucination ASH{+} framework is designed to enhance the stylistic diversity of the training data. 
Building on the idea from prior work \cite{park2019SPADE,Yue_2019_ICCV}, we condition the stylization process of the training images (source domain images) with semantic information.

Specifically, our approach employs a dual-stage manipulation of hallucination strength. 
The first stage involves varying the stylistic information of the style features by modulating them with the semantic information. 
In the second stage, we adjust the stylization strength of the images based on their semantic content when generating new images by a decoder.

Our new method surpasses our earlier proposed approach ASH \cite{Tjio_2022_WACV}. 
ASH relied solely on the first stage which adjusts the stylistic diversity of the stylized source domain images. 
It used a predefined image-level value to control the overall stylization strength across all source domain images. 
In contrast, ASH{+} introduces an additional stage which allows for finer-grained control of stylization strength based on the semantic details of each image. 
This additional stage improves performance and increases adaptability to the class-wise differences in the training data.

As illustrated in \textbf{Figure \ref{figure:workflow_fig}}, our designed ASH{+} framework is composed of three main components: a segmentation module $G$, a dual-stage manipulation of hallucination strength layer (\textit{dFT}), and a pre-trained VGG encoder/decoder. 
The segmentation module $G$ is responsible for performing the semantic segmentation task, while the dual-stage manipulation of hallucination strength layer is designed to enhance the diversity of the training data through hallucination modulated by semantic information. The pre-trained VGG encoder/decoder is used for feature extraction and reconstruction.


 \textbf{VGG Encoder} In our approach, we utilize a ImageNet \cite{ImageNet} pretrained VGG19 encoder, denoted as $Enc$, to extract both the style features $\boldsymbol{f_{style}}$ and the content features $\boldsymbol{f_{src}}$ from the style image $\boldsymbol{X_{style}}$ and the source image $\boldsymbol{X_{s}}$, respectively. 
 The VGG19 encoder is used to map the images from their pixel space to the latent space, where representations of the images in different levels are captured.

\textbf{Segmentation module $G$}
We obtain the one-hot predicted {class probabilities} $P_{ij}^{k}$ from the source domain images $\boldsymbol{X_{s}}$ using the segmentation network $G$.
We then generate {$\boldsymbol{\phi}$ by mapping $P_{ij}^{k}$ with the embedding layer to the common latent space shared by the style features $\boldsymbol{f_{style}}$ and source features $\boldsymbol{f_{src}}$.}
\small
\begin{equation}
\boldsymbol{\phi}= \text{Embedding layer}(P_{ij}^{1},P_{ij}^{2},...,P_{ij}^{k}, ...,P_{ij}^{K}),
\end{equation}
\normalsize
where $K$ refers to the number of classes and \textit{ij} refers to the spatial position in the image.
We use $\phi$ to generate manipulation parameters that control the hallucination process in dual stages. \footnote{While it is possible to obtain the semantic information at several different scales\cite{lin2020scn}, we simply map the pixel-level predictions (from the classifier layer) to a single latent space encoding.
This reduces the computational complexity of computing the affine transformations for the different scales before merging the transformed features.}

$dFT$ ( \textbf{d}ual-stage \textbf{F}eature \textbf{T}ransform) \textbf{layer} 
To control the hallucination strength in the first stage, the \textit{dFT} layer generates the style-content re-balancing factor $\boldsymbol{\alpha}$, the class-wise scaling $\boldsymbol{\gamma}$, and the shifting $\boldsymbol{\beta}$ coefficients via the following equation:
\small
\begin{equation}
\boldsymbol{\gamma},\boldsymbol{\beta},\boldsymbol{\alpha}=dFT(\boldsymbol{\phi}),
\label{eqn:classwise_scaleshift}
\end{equation}
\normalsize
where $\boldsymbol{\gamma, \beta}$ and $\boldsymbol{\alpha}$ share the same spatial and channel dimensions as the style/content features. 

Given the class-wise scaling $\boldsymbol{\gamma}$, the shifting $\boldsymbol{\beta}$, the \textit{dFT} {layer} linearly transforms the style features $\boldsymbol{f_{style}}$  via the following equation {in the first stage of hallucination strength manipulation}:
\small
\begin{equation}
\boldsymbol{f_{style}^{'}}=\boldsymbol{\gamma}(\boldsymbol{z}.\boldsymbol{f_{style}}+1) + \boldsymbol{\beta},
\label{eqn:perturbed style features}
\end{equation}
\normalsize
where $\boldsymbol{z}$ denotes a randomly generated {unit noise tensor that is orthogonal to $\boldsymbol{f_{style}}$}\footnote{Inspired by previous work\cite{Noise_training_generalization,bishop_noise,Tjio_2022_WACV}, which demonstrated that adding noise to the inputs can improve model generalizability, we introduced noise $\boldsymbol{z}$ orthogonal to the style features. 
The motivation behind this design choice was to increase style diversity in the style features.}.



{For the first-stage manipulation of hallucination strength, we control the stylization of image pixels depending on their class-wise properties through affine transformations of the style features.
The affine transformations in feature space increase stylistic diversity and are also adversarially updated, allowing us to generate increasingly challenging examples for training.}



\textbf{VGG Decoder}
We generate the stylized source domain images $\boldsymbol{X_{stylized}}$ with the following equation:
\small 
\begin{equation}
\boldsymbol{X_{stylized}} =Dec((\boldsymbol{\sigma^\prime_{1}} +\boldsymbol{\alpha})\boldsymbol{f_{src}}+(\boldsymbol{\sigma^\prime_{2}} -\boldsymbol{\alpha)}\text{AdaIN}(\boldsymbol{f_{src}},\boldsymbol{f_{style}^{'}})),
\label{eqn:stylizedsrc}
\end{equation} 
\normalsize
where $Dec$ is a ImageNet\cite{ImageNet} pretrained VGG19 decoder, AdaIN is the adaptive instance normalization equation defined in \textbf{Equation \ref{eqn:adain}}.
{We derive }$ \boldsymbol{\sigma^\prime_{1}}$ and $ \boldsymbol{\sigma^\prime_{2}}$ {from the two predefined hyper-parameters $\sigma_{1}$ and $\sigma_{2}$ by scalar multiplication with a unit tensor.
$Dec$ maps the features from the latent space back to pixel space. 

Recent work \cite{ahn2023cuda} demonstrated the effectiveness of a class-wise approach to determine augmentation strength.
In a similar vein, we leverage the semantic information in the second stage of our semantic manipulation of hallucination strength to generate the style-content balancing weight $\boldsymbol{\alpha}$. 
$\boldsymbol{\alpha}$ controls the effective stylization strength by balancing the element-wise proportions of the original source features $\boldsymbol{f_{src}}$ with the stylized source features.
{After merging the features, we reconstruct the stylized source domain images (Equation \ref{eqn:stylizedsrc}).}

{In summary, we employ a dual-stage manipulation of hallucination strength to increase stylistic diversity and adaptively adjust stylization strength.
We then include the stylized images during training to improve model generalizability.
}



\subsection{Framework Learning}
\label{subsection:trainingdetails}
{We train the two sub-modules in our framework, namely the segmentation network \textit{G} and the \textit{dFT} layer.}

\textbf{Optimization for $G$}
The segmentation network $G$ \cite{Luo_2019_CVPR} is trained to minimize segmentation loss $\mathcal{L}_{seg}$ and the pixel-wise consistency loss $\mathcal{L}_{cont}$. 
  
Segmentation loss $\mathcal{L}_{seg}(G,\boldsymbol{X},\boldsymbol{Y})$ is derived from computing the cross entropy loss for the segmentation output\cite{Luo_2019_CVPR}:
\small
\begin{equation}\
\begin{aligned}
&\mathcal{L}_{seg}(G,\boldsymbol{X},\boldsymbol{Y}) = \sum_{i=1}^{H \times W}\sum_{c=1}^{C} -Y_{i,c}log(G(X_{i,c})),
\label{eqn:cross_ent_eqn}
\end{aligned}
\end{equation}
\normalsize
where $G(X_{i,c})$ refers to the predicted probability of class $c$ on the \textit{i}th pixel. $Y_{i,c}$ is the known ground truth probability for class $c$ on the \textit{i}th pixel, where $Y_{i,c}=1$ if the pixel belongs to the class \textit{c} and $Y_{i,c}=0$ if otherwise.

This pixel-wise consistency loss is given by the following equation:
\small
\begin{equation}\
\begin{aligned}
&\mathcal{L}_{cont}(\boldsymbol{G, X_{i}^{stylized}},\boldsymbol{X_{i}^{s}}) = \\&\frac{1}{H \times W} \sum_{i=1}^{H \times W}KL(G(\boldsymbol{X_{i}^{stylized}})||G(\boldsymbol{X_{i}^{s}})),
\label{eqn:consistency}
\end{aligned}
\end{equation}
\normalsize
where $G(X_{i}^{stylized})$
represents the class probabilities of the \textit{i}th pixel in the segmentation output of the stylized source domain data, $G(X_{i}^{s})$ represents the class probabilities of the \textit{i}th pixel in the segmentation output of the original source domain data, $KL(·.)$ is the Kullback-Leibler divergence between two probabilities.
{The pixel-wise consistency loss function is based on the formulation \cite{8954081}, which computes the Kullback-Leibler divergence between the pixel-wise segmentation output of the original source domain image and the segmentation output from the stylized source domain image.}

In our previous work \cite{Tjio_2022_WACV}, we utilized a discriminator to enforce domain alignment and enhance domain generalization. 
However, in the current work, we opted for a different approach by replacing the discriminator with the pixel-wise consistency loss (Equation \ref{eqn:consistency}). 
The primary motivation behind this change was to reduce the training cost and complexity.
Despite removing the discriminator from our workflow, we still achieved comparable performance to our earlier reported results \cite{Tjio_2022_WACV}.

\textbf{Optimization for $dFT$}
We optimize the \textit{dFT} layer by maximizing pixel-wise consistency loss $\mathcal{L}_{cont}$. 
This encourages the $dFT$ layer to create challenging training data for the segmentation network $G$. 
We minimize content loss $\mathcal{L}_{c}$ (\textbf{Equation \ref{eqn:contentloss}}) to preserve the semantic information present in the source features.
We also minimize style loss $\mathcal{L}_{s}$ (\textbf{Equation \ref{eqn:styleloss}}) from the perturbed style features $\boldsymbol{f_{style}^{'}}$ to maximise the stylistic diversity of the generated images.
We compute the loss for $dFT$ with the following equation: 
\small
\begin{equation}
\begin{aligned}
&\mathcal{L}_{ASH+}(G,\boldsymbol{f_{src}},\boldsymbol{f_{style}^{'}},\boldsymbol{X_{stylized}},\boldsymbol{X_{s}}) = \\ &-\mathcal{L}_{cont}(\boldsymbol{G,X_{stylized}},\boldsymbol{X_{s}})\\ &+ \mathcal{L}_{c}(\boldsymbol{f_{src}},\text{AdaIN}(\boldsymbol{f_{src}},\boldsymbol{f_{style}^{'}}))\\ &+ \mathcal{L}_{s}(\boldsymbol{f_{style}^{'}},\text{AdaIN}(\boldsymbol{f_{src}},\boldsymbol{f_{style}^{'}}))\\ &- \mathcal{L}_{s}(\boldsymbol{f_{src}},\text{AdaIN}(\boldsymbol{f_{src}},\boldsymbol{f_{style}^{'}}))
\label{eqn:ASHloss}
\end{aligned}
\end{equation}
\normalsize

In addition, we use the formula for content loss and style loss as defined in \cite{huang2017adain}. 
{Here, we minimize the amount of style information retained from the source features $\boldsymbol{f_{src}}$ while simultaneously preserving semantic information from the source domain in the merged features $\text{AdaIN}(\boldsymbol{f_{src}},\boldsymbol{f_{style}^{'}})$.
The objective is to avoid retaining the domain variant features from the source domain while maximising the semantic information (\ie domain invariant information) in the stylized images.}

Content loss $\mathcal{L}_{c}$ is derived from the $L2$ norm between the original source domain features and the merged style-source features.  
\small
\begin{equation}
\mathcal{L}_{c}=\| \boldsymbol{f_{src}}-\text{AdaIN}(\boldsymbol{f_{src}},\boldsymbol{f_{style}^{'}}) \|_{2},
\label{eqn:contentloss}
\end{equation}
\normalsize
Style loss $\mathcal{L}_{s}$ is computed from the mean and standard deviation of the features extracted from each of the \textit{L} convolution layers in the VGG19 encoder.
\small
\begin{equation}
\begin{aligned}
\mathcal{L}_{s} =\sum_{i=1}^{L}(&\| \mu(\boldsymbol{f_{src}})-\mu(\text{AdaIN}(\boldsymbol{f_{src}},\boldsymbol{f_{style}^{'}}))\| \\
& + \| \sigma(\boldsymbol{f_{src}})-\sigma(\text{AdaIN}(\boldsymbol{f_{src}},\boldsymbol{f_{style}^{'}}))\| ),
\end{aligned}
\label{eqn:styleloss}
\end{equation}
\normalsize 

The training workflow is summarized in \textbf{Algorithm \ref{alg:workflow_alg}}. 
It is noted that the weights for the pretrained encoder and decoder that are used during stylization are not updated during training. After training, we only need the segmentation network $G$ for evaluation and the ASH{+} module is not required.

\section{Experiments}
\subsection{Datasets}
\label{subsection:datasets}
We assess the performance of our proposed method on the following publicly available datasets:
\begin{table*}[ht]
\begin{center}
 \normalsize
 \resizebox{\linewidth}{!}{%
 \begin{tabular}{p{1.5cm}|cccccc|ccccc|cc }
\toprule
\multicolumn{4}{l}\textbf{New York} &\multicolumn{7}{l}\textbf{Old European Town} \\
 \midrule
 Source Domain & Method & Dawn & Fog & Night & Spring & Winter & Dawn & Fog & Night & Spring & Winter & Avg. mIoU \textuparrow & Rel. Diff. (\%)  \textuparrow  \\
  \midrule
 & ERM\cite{koltchinskii2011oracle} & 
 27.80 &2.73& 0.93& 6.80& 1.65& 52.78& 31.37& 15.86& 33.78& 13.35& 18.70 & 48.48\\
 & M-ADA \cite{qiaoCVPR20learning} & 29.10& 	4.43& 	4.75& 	14.13 &	4.97& 	54.28& 	36.04& 	23.19& 	37.53& 	14.87& 	22.33 & 38.48 \\
 Highway   & PDEN \cite{Li_2021_CVPR} & 30.63 & 21.74  & 16.76  & 26.10 & 19.91& 54.93 
&47.55  & 36.97 &43.98  &23.83  &32.24 & 11.18\\
(Dawn)& Source-only & 24.50 & 2.90 & 0.40 & 1.80 & 0.06 & 53.90 &33.50 &18.00 &30.90 &12.80 & 17.90 & 50.69\\
&ASH\cite{Tjio_2022_WACV} & 16.70 & 16.00 & 11.70 & 14.20 & 12.80&23.80 &20.00 &13.20 &20.00 & 12.40 & 16.01
 & 55.90 \\
&ASH{+} & \textbf{30.70}& \textbf{30.00}& \textbf{27.30}& 
\textbf{31.20}& \textbf{27.30}& \textbf{52.30}&
\textbf{50.10}& \textbf{43.50}& \textbf{45.60}&
\textbf{25.00}& \textbf{36.30} & - \\ 
\midrule
 & ERM\cite{koltchinskii2011oracle} & 
17.24&34.80&12.36&26.38&11.81&33.73&55.03&26.19& 41.74&12.32&27.16 & 21.37\\
& M-ADA \cite{qiaoCVPR20learning} &21.74&	32.00&	9.74&	26.40	&13.28&	42.79&	56.60&	31.79&	42.77&	12.85&	29.00 & 16.04 \\
Highway  & PDEN \cite{Li_2021_CVPR}& \textbf{25.61} & 35.16 & 17.05&\textbf{32.45}& {21.03} &\textbf{45.67}& \textbf{54.91} &\textbf{37.38}& \textbf{48.29}& 20.80& 33.83& 2.06 \\
(Fog)&  Source-only & 
18.80 & 34.70 & 12.50 & 26.30 & 13.20 & 36.10 & 54.00 & 28.80 & 42.00 & 14.10 & 28.05 & 18.79\\
&ASH\cite{Tjio_2022_WACV}& 11.70 & 20.30 & 11.80 & 19.30 & 14.00 & 13.70 & 32.00 & 21.20 & 24.90 & 16.60 &
 18.55 & 46.30 \\
&ASH{+}& 25.60& \textbf{36.20}& \textbf{19.40}& 31.80& \textbf{27.30}& 42.30& 54.70& 37.00& 44.00& \textbf{27.10}&
\textbf{34.54} & - \\
\midrule
 & ERM \cite{koltchinskii2011oracle} & 26.75 & 26.41& 18.22 &32.89& 24.60& 51.72 &51.85& 35.65& 54.00 &28.13& 35.02 & 15.39\\
 & M-ADA \cite{qiaoCVPR20learning} & 29.70 &	31.03&	22.22&	38.19&	28.29&	53.57&	51.83&	38.98&	55.63&	25.29&	37.47 & 9.47\\
Highway & PDEN\cite{Li_2021_CVPR} &28.17 & 27.67 &27.53 &34.30 &28.85 &\textbf{53.75}& 51.53& 46.87& 55.63& \textbf{30.61} &38.49 & 7.01\\
(Spring)& Source-only & 27.10 & 27.70 & 15.90 &
31.60 & 25.00 & 52.50 & 50.50 & 33.50 & 54.90 & 28.30 & 34.70 & 16.16\\
&ASH \cite{Tjio_2022_WACV}& 8.30 & 12.40 & 5.50 & 15.30 & 10.40 & 23.20& 30.80 & 16.40 & 31.70 & 20.90 
 & 17.49 & 57.74 \\
&ASH{+}& \textbf{34.40}& \textbf{36.50}& \textbf{32.60}& \textbf{40.60}& \textbf{29.00}& 52.60& \textbf{54.80}& \textbf{49.40}& \textbf{55.80}& 28.20&
\textbf{41.39} & -\\
\bottomrule
\end{tabular}}
\end{center}
\caption{Quantitative semantic segmentation results on SYNTHIA\cite{RosCVPR16}. The models are trained using a single-source domain and evaluated on several unseen target domains. We report the averaged mean Intersection Over Union (Avg. mIoU) and also show the visual results in \textbf{Figure \ref{figure:qualitative_cityscapes }}. 
{Relative difference (\%) indicates the performance improvements with our approach compared to the comparison method. }} 
\label{tab:multi_sythia_proof}
\end{table*}

\begin{figure*}
\begin{center}
\includegraphics[width=1.0\textwidth,keepaspectratio]{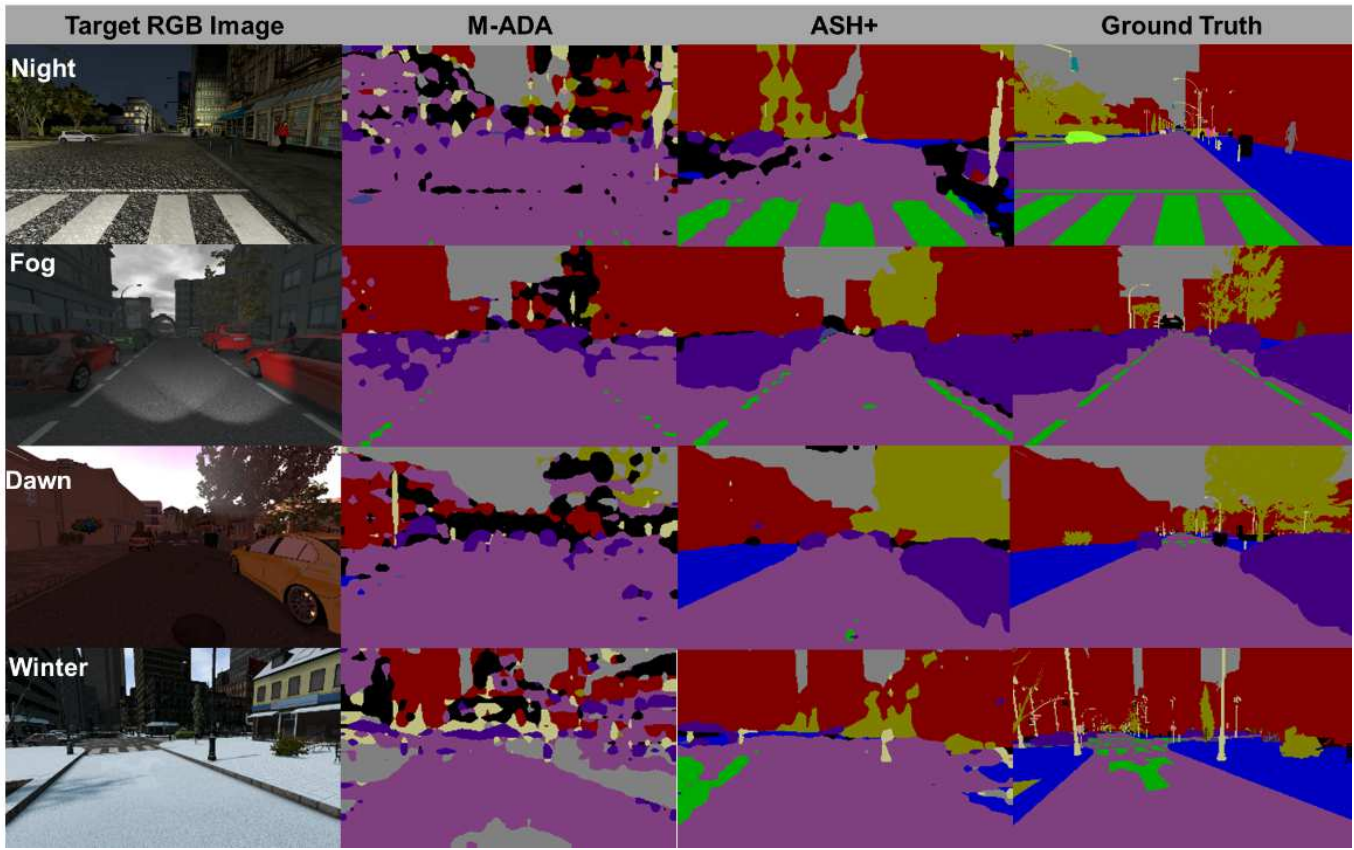}
\end{center}
\caption{Qualitative comparison of segmentation output for SYNTHIA\cite{RosCVPR16}. For each image, we show the corresponding results for ``M-ADA" \cite{qiaoCVPR20learning}, our proposed method ASH{+} on the original image and the corresponding ground truth. The model used was trained on Highway (Spring) and the results on the unseen weather conditions (Night, Fog, Dawn and Winter) are shown.}
\label{figure:sythia comparison}
\end{figure*}

\begin{table*}[t]
 \begin{center}
 \normalsize
 \resizebox{\linewidth}{!}{%
 \begin{tabular}{p{2cm}| c|abababababababababab|cc}
  \toprule
  \multicolumn{24}{c}{\textbf{GTA5 $\rightarrow$ Cityscapes}} \\
  \toprule & \rotatebox{90}{Year} & \rotatebox{90}{Arch.}
   & \rotatebox{90}{road} & \rotatebox{90}{side.} & \rotatebox{90}{buil.} & \rotatebox{90}{wall} & \rotatebox{90}{fence} & \rotatebox{90}{pole} & \rotatebox{90}{light} & \rotatebox{90}{sign} & \rotatebox{90}{vege.} & \rotatebox{90}{terr.} & \rotatebox{90}{sky} & \rotatebox{90}{pers.} & \rotatebox{90}{rider} & \rotatebox{90}{car} & \rotatebox{90}{truck}& \rotatebox{90}{bus} & \rotatebox{90}{train} & \rotatebox{90}{motor} & \rotatebox{90}{bike} &  \rotatebox{90}{\textbf{mIoU}} & \rotatebox{90}{Rel. (\%) Diff.} \\ 
   \toprule
   \midrule
 Source only & - &R & 75.8 & 16.8 & 77.2 & 12.5 & 21.0 & 25.5 & 30.1 & 20.1 & 81.3 & 24.6 & 70.3 & 53.8 & 26.4 & 49.9 & 17.2 & 25.9 & 6.5 & 25.3 & 36.0 & 36.6 & - \\
  Fully supervised & -&R &97.9 &81.3 &90.3 &48.8 &47.4 &49.6 &57.9 &67.3 &91.9 &69.4 &94.2 &79.8 &59.8 &93.7 &56.5 &67.5 &57.5 &57.7 &68.8 &70.4 & - \\
 \midrule
 \multicolumn{24}{c}{\normalsize{Domain Generalization}}\\
\midrule
 Advent \cite{Vu_2019_CVPR}& 2019 &R &83.00 &1.80 &72.00 &8.20 &3.60 &16.20 &22.90 &9.80 &79.30 &17.10 &75.70 &35.10 &15.80 &70.90 &30.90 &35.30 &0.00 &16.40 &24.90 &32.60 &20.10\\
 MaxSquare \cite{Chen_2019_ICCV}& 2019 &R &76.80 &14.20 &77.00 &18.80 &14.10 &14.50 &30.30 &18.00 &79.30 &11.70 &70.50 &53.00 &24.20 &68.70 &25.30 &14.00 &1.30 &20.60 &25.50 &34.60&15.20\\
 CLAN \cite{Luo_2019_CVPR}& 2019 &R &  87.20 &20.10 &77.90 &25.60 &19.70 &23.00 &30.40 &22.50 &76.80 &25.20 &76.20 &55.10 &28.10 &82.70 &30.70 &36.90 &0.80 &26.00 &17.10 &40.10 &1.72 \\ 
 ASM\cite{DBLP:conf/nips/LuoLGY020}& 2020 &R & 56.20 &0.00 &7.00 &0.60 &1.00 &0.30 &0.70 &0.60 &13.80 &0.10 &0.01 &0.08 &0.04 &1.20 &0.50 &0.70 &0.20 &0.00 &0.00 & 4.40 & 89.22\\
    \midrule
Domain Rand.\cite{Yue_2019_ICCV}& 2019 &R & - &- &- &- &- &- &- &- &- &- &- &- &- &- &- &- &- &- &- &42.53 & -4.24  \\
ASH{+}&2023& R & 85.06 &
35.11& 77.52& 24.93& 16.96& 24.02& 27.44&
23.68& 79.93& 28.64& 69.59& 57.49& 29.16&
80.28& 26.41& 39.10& 4.59& 19.17& 26.06&40.80\\
SOMAN\cite{Kundu_2021_ICCV}*& 2021& R & 85.84 & 36.53 & 81.79 &  24.77  & 22.38  & 29.03  & 30.25  & 21.04  & 83.61   & 36.09   & 79.89  & 56.72   & 26.56 & 83.73  & 36.29   &  41.37 & 0.04  & 20.56   & 22.89 & \textbf{43.13} & \textbf{-5.71}\\
\bottomrule
\end{tabular}}
\end{center}
\caption{Segmentation performance of Deeplab-v2 with ResNet-101 backbone for GTA5\cite{Richter_2016_ECCV}\textrightarrow{} Cityscapes\cite{Cordts2016Cityscapes}. 
(*: We evaluated this work at a single scale to maintain consistency with all other comparisons. ) 
While our approach performs less well than the DRPC \cite{Yue_2019_ICCV} or SOMAN \cite{Kundu_2021_ICCV}, it must also be noted that these methods utilize a much larger batch size and are computationally more costly.
{Relative difference (\%) indicates the performance improvements with our approach compared to the comparison method.}}
\label{tab:gta-cityscapes-paper-proof}
\end{table*}

\begin{table*}[h]
\centering
\scriptsize
\resizebox{\linewidth}{!}{%
\begin{tabular}{p{2cm}|c|ababababababab|c|cc}%
\toprule
  \multicolumn{18}{c}{\textbf{Synthia $\rightarrow$ Cityscapes}} \\
  \midrule &\rotatebox{90}{Year} & \rotatebox{90}{Arch.}
   & \rotatebox{90}{road} & \rotatebox{90}{side.} & \rotatebox{90}{buil.} & \rotatebox{90}{light} & \rotatebox{90}{sign} & \rotatebox{90}{vege.} & \rotatebox{90}{sky} & \rotatebox{90}{pers.} & \rotatebox{90}{rider} & \rotatebox{90}{car} & \rotatebox{90}{bus} & \rotatebox{90}{motor} & \rotatebox{90}{bike} & \rotatebox{90}{\textbf{mIoU}}& \rotatebox{90}{\textbf{mIoU16}} & \rotatebox{90}{Rel. (\%) Diff} \\ 
   \midrule
   \midrule
Source only &-& R & 47.10 & 19.50 & 68.90 & 9.10 & 9.30 & 75.10 & 79.10 & 52.50 & 20.30 & 43.00 & 20.70 & 9.40 & 29.30 & 37.17& 32.38  & -  \\
Fully supervised &-& R &95.10 &72.90 &87.30 &46.70 &57.20 &87.10 &92.10&74.20 &35.00 &92.10 &49.30 &53.20 &68.80 &70.10&  - & - \\
 \midrule
 \multicolumn{17}{c}{\scriptsize{Domain Generalization}}\\
   \midrule
 Advent \cite{Vu_2019_CVPR}& 2019 &R  & 72.30 &30.70 &65.20 &4.10 &5.40 &58.20 &77.20 &50.40 &10.10 &70.00 &13.20 &4.00 &27.90 &37.60& 31.80 & 20.10\\
 MaxSquare \cite{Chen_2019_ICCV} & 2019 &R & 57.80 &23.19 &73.63  &8.37 &11.66 &73.84 &81.92 &56.68  &20.73  &52.18  &14.71  &8.37 &39.18& 40.17 &34.96 & 12.16 \\
 CLAN \cite{Luo_2019_CVPR} & 2019 &R & 63.90 &25.90 &72.10 &14.30 &12.00 &72.50 &78.70 &52.70 &14.50 &62.20 &25.10 &10.40 &26.50 &40.90& 34.90 & 12.31\\ 
 ASM \cite{DBLP:conf/nips/LuoLGY020} & 2020 &R  &75.40 & 18.50 &66.60 &0.10 &0.80 &67.00 &77.80 &15.60 &0.50 &11.40 &1.30 &0.03 &0.20 & 25.80 & 21.60 & 45.73 \\
 \midrule
Domain Rand.\cite{Yue_2019_ICCV} & 2019 &R & - &- &- &- &- &- &- &- &- &- &- &- &- &- &37.58 & 5.58 \\
SoMAN\cite{Kundu_2021_ICCV} & 2021 &R & 87.20 & 34.95 & 77.25 &15.61&18.25& 74.98 & 78.26 & 52.59 & 21.47 & 76.17 & 29.31 & 13.74 & 26.79 &46.7&\textbf{40.10} & -0.75 \\
ASH{+} & 2023 & R & 76.29 &30.64 & 75.54&  17.44& 17.55& 73.25&  77.17& 53.43& 22.18& 75.75&  31.80& 11.89& 37.31& 46.17 & 39.80 & - \\
\bottomrule
 \end{tabular}}
 \caption{
   Segmentation performance of Deeplab-v2 with ResNet-101 backbone for Synthia\cite{RosCVPR16}\textrightarrow{} Cityscapes\cite{Cordts2016Cityscapes}.
   {Relative difference (\%) indicates the performance improvements with our approach compared to the comparison method.}
  } 
   \label{tab:synthia-cityscapes-proof}
\end{table*}

\begin{itemize}
    \item \textbf{SYNTHIA}\cite{RosCVPR16} is a synthetic semantic segmentation dataset of urban scenes with varying weather conditions and illumination levels, across 3 different environments (Highway, New York-like and Old European Town). 
    We use the 13 semantic categories in the dataset. 
    Similar to previous work \cite{qiaoCVPR20learning}, we use images from Dawn/Spring/Fog in the Highway environment as the source domain. 
    We train the models separately under each of the different weather conditions and evaluate them on the New York-like and Old European Town subsets.
    \item \textbf{GTA5}\cite{Richter_2016_ECCV} is a synthetic semantic segmentation dataset with 24,966 densely annotated images with resolution $1914 \times 1052$ pixels, and has 19 categories that are compatible with the Cityscapes\cite{Cordts2016Cityscapes} dataset.
    \item \textbf{Synthia}\cite{RosCVPR16} refers to the  SYNTHIA-RAND-CITYSCAPES subset from the publicly available database for semantic segmentation. 
    It has 9,400 densely annotated images with resolution $1280 \times 760$ pixels and has 16 categories that are compatible with the Cityscapes\cite{Cordts2016Cityscapes} dataset.
    \item \textbf{Cityscapes}\cite{Cordts2016Cityscapes} is a real-world driving dataset with densely annotated images of resolution 2048 $\times$ 1024 pixels. 
    We use the validation split of 500 densely annotated images to evaluate model performance.
\end{itemize}

We use 50,000 images from the ImageNet \cite{ImageNet} validation data (ILSVRC2011) as a source of style images. 
We use the mean Intersection over Union (mIoU), which is a widely used performance metric that quantifies the overlap between the ground truth labels and the predicted output.

\subsection{Implementation Details}
\textbf{SYNTHIA} For the SYNTHIA dataset, we follow the implementation proposed by Qiao \etal \cite{qiaoCVPR20learning}. The training and testing images are sourced from the left-front camera and are resized to a resolution of 192 $\times$ 320 pixels.
For the model architecture, we employ a Fully Convolutional Network (FCN) \cite{7298965} with a ResNet-50 backbone \cite{He2015} pretrained on ImageNet \cite{ILSVRC15}. 
The Adam optimizer \cite{adam} is utilized during training.

Our training process consists of two phases. In the first phase, we train the network on the source domain data for 50 epochs. 
Subsequently, we proceed to the joint training phase with the $dFT$ layer for a maximum of 10 epochs. 
During joint training, we use a batch size of 8 images and conduct the training on a single 12 GB Titan X GPU. 
Our implementation is based on PyTorch \cite{NEURIPS2019_9015}.

\textbf{GTA5/Synthia\textrightarrow Cityscapes} 
For the tasks GTA5$\rightarrow$ Cityscapes and Synthia\textrightarrow Cityscapes, we implement our approach based on the implementation of Luo \etal \cite{Luo_2019_CVPR}. 
The GTA5 images are resized to $1280 \times 720$ pixels, while the Synthia images are resized to $1280 \times 760$ pixels.

The segmentation network used in this task is the Deeplab-v2 network \cite{7913730} with a ResNet-101 \cite{He2015} backbone, pretrained on the ImageNet dataset \cite{ILSVRC15}. 
We initialize the learning rate to $2.5$$\times$$10^{-4}$ and use stochastic gradient descent (SGD) with a momentum of $0.9$ to optimize both the segmentation network and the $dFT$ layer.
The training process involves $100,000$ iterations ($Iter_{num}$) from Algorithm \ref{alg:workflow_alg} on a single 16 GB Quadro RTX 5000 GPU. 

\subsection{Comparison with Competing methods}

\textbf{SYNTHIA} We conduct a comprehensive evaluation of the generalizability of our proposed ASH{+} framework across different weather conditions and environments in the SYNTHIA dataset. 
While there might be a larger domain gap between synthetic and real-world images (\ie GTA5/Synthia\textrightarrow Cityscapes), SYNTHIA has multiple unseen target domain domains (more than 2) while (GTA5\slash Synthia\textrightarrow Cityscapes) only has a single unseen target domain. To benchmark our method, we compare it with several state-of-the-art approaches, including M-ADA\cite{qiaoCVPR20learning} and PDEN\cite{Li_2021_CVPR}, which also leverage data transformations to increase training data diversity. 
Additionally, we include a representative work, Empirical Risk Minimization (ERM)\cite{koltchinskii2011oracle}, in the comparison. 
By evaluating our method alongside these established approaches, we can effectively gauge its performance and demonstrate its effectiveness in addressing the single-source domain generalization problem.

ERM\cite{koltchinskii2011oracle} refers to the method where the model is derived from minimizing the average loss (\ie cross entropy loss) over the training data and is most similar to that of the source-only approach.
 Qiao \etal\cite{qiaoCVPR20learning} (M-ADA) apply a Wasserstein Autoencoder to generate additional data via transformations of the source data and use a relaxation loss function to maximise the difference between the transformed data and original source data. 
The task model is then updated to minimize loss on the original source data (meta-train) and the transformed data (meta-test).

Li \etal\cite{Li_2021_CVPR} (PDEN) progressively generate data via multiple generators that estimate the style and colour information present in the unseen target domain data, followed by contrastive learning to help the model to learn good domain-invariant representations for the different classes.

{The quantitative results are shown in \textbf{Table \ref{tab:multi_sythia_proof}} and a qualitative comparison is shown in \textbf{\textbf{Figure} \ref{figure:sythia comparison}}. 
For all three settings (Highway-Dawn\slash Fog\slash Spring), ASH{+} yields superior performance compared to recent state-of-the-art work \cite{Li_2021_CVPR,qiaoCVPR20learning}, further validating the effectiveness of ASH{+} for domain generalization tasks.
Unlike Qiao \etal\cite{qiaoCVPR20learning}(M-ADA), which uses a Wasserstein Autoencoder ($10^{9}$ 
trainable parameters), our approach uses 100x fewer parameters.
Similar to our method, PDEN\cite{Li_2021_CVPR} adversarially generates training data to improve generalizability. 
However, like M-ADA\cite{qiaoCVPR20learning}, {the generation process} is not conditioned with semantic information
which could have reduced performance for the challenging minority classes.}

{{Additionally, {our proposed} style-content balancing weight $\boldsymbol{\alpha}$ builds upon our previous work ASH \cite{Tjio_2022_WACV} and demonstrates improved performance.
Though ASH \cite{Tjio_2022_WACV} generates images that are conditioned with semantic information, the overall stylization strength is determined via a global predefined value.
It is also possible that the global predefined value is poorly suited for the SYNTHIA \cite{RosCVPR16} dataset, resulting in poorer performance (\textbf{Table \ref{tab:multi_sythia_proof}}).}}

\textbf{GTA5/Synthia\textrightarrow Cityscapes} We compare our method with state-of-the-art work DRPC \cite{Yue_2019_ICCV} and SOMAN \cite{Kundu_2021_ICCV}. 
{
We also implement representative unsupervised domain adaptation work \cite{Vu_2019_CVPR, Chen_2019_ICCV, Luo_2019_CVPR,DBLP:conf/nips/LuoLGY020} to provide additional comparisons.
DRPC\cite{Yue_2019_ICCV} generates multiple stylized images from a single-source domain image {per training iteration} and trains the model to maximise consistency within the batch.} 
SOMAN \cite{Kundu_2021_ICCV} involves training a segmentation model containing a single global classifier head and 5 leave-one-out classifier heads with 5 groups ('domains`) of differently augmented source domain data.

For the representative unsupervised domain adaptation work \cite{Vu_2019_CVPR, Chen_2019_ICCV, Luo_2019_CVPR,DBLP:conf/nips/LuoLGY020}, we apply the style transfer approach\cite{huang2017adain} to generate pseudo target domain images for training.
Advent \cite{Vu_2019_CVPR} improves performance on the unlabeled target domain data by minimizing prediction entropy on the target domain data while simultaneously improving performance on the labeled source domain data.
Maximum squares loss\cite{Chen_2019_ICCV} minimizes the square of the probability predictions for the unlabeled target domain data to improve performance for the minority classes during adaptation.
CLAN \cite{Luo_2019_CVPR} applies a co-training approach to learn domain invariant representations for the source and target domain data.
Finally, ASM\cite{DBLP:conf/nips/LuoLGY020} adversarially updates the style information from a single target-domain image, and uses the updated style information to generate increasingly challenging pseudo target domain images for adaptation.
\begin{figure*}
\centering
\includegraphics[width=1.0\textwidth,keepaspectratio]{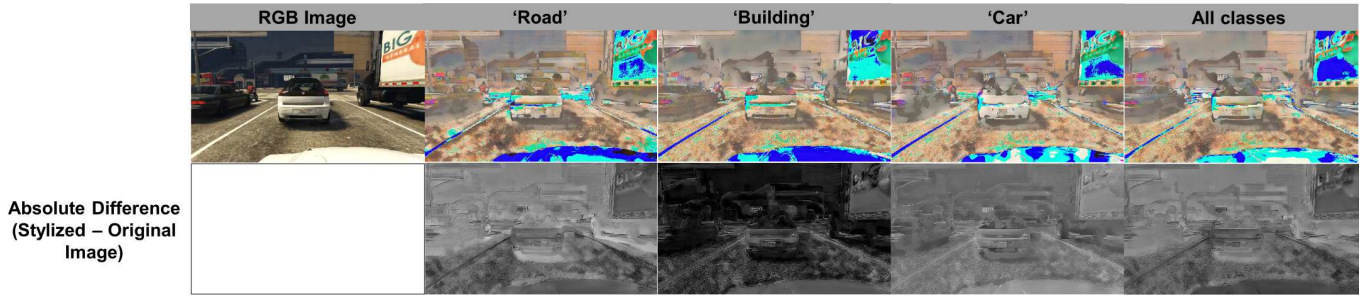}
\caption{Visualization of absolute stylization differences across classes for a single test image. For the classes listed (\ie road/building/car), we exclude all other 18 classes in the segmentation output and stylize the source domain image only with the prediction outputs for that specified class. The absolute difference between the stylized image and the original source domain image is shown in the bottom row. The stylized image with semantic information from all classes is indicated as "all classes" in the rightmost column.}
\label{figure:diff_plot}
\end{figure*}
\begin{figure*}
\begin{center}
\includegraphics[width=1.0\textwidth,keepaspectratio]{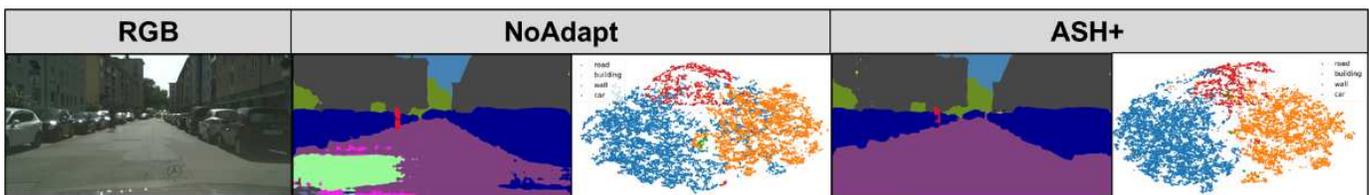}
\end{center}
 \vspace{-0.5cm}
\caption{t-SNE plots for NoAdapt (model trained on source domain data without any data augmentation or domain generalization approaches) and our approach, ASH{+}. 
{There is considerable overlap between the distributions 'road', 'building', and 'car' for NoAdapt.
In contrast, our method shows better separation for the different class distributions.}
A representative segmentation output from each model is included. }
\label{figure:tsne plot}
\end{figure*}



The quantitative results for GTA5/Synthia$\rightarrow$ Cityscapes are shown in \textbf{Table} \ref{tab:gta-cityscapes-paper-proof} and \textbf{Table}  \ref{tab:synthia-cityscapes-proof}, respectively. 
We  show the qualitative results for Synthia\textrightarrow Cityscapes in \textbf{Figure \ref{figure:qualitative_cityscapes }}.
For Synthia\textrightarrow Cityscapes, our approach demonstrates improved performance compared to DRPC \cite{Yue_2019_ICCV} and similar performance with SOMAN \cite{Kundu_2021_ICCV}.
This is significant because our approach uses only a single stylization approach and only requires a batch size of 1 (1 source domain image and 1 stylized source domain image).
In contrast, DRPC \cite{Yue_2019_ICCV} requires a large batch size of stylized training images (n=15).
While SOMAN \cite{Kundu_2021_ICCV} uses a smaller batch size (n=5), it requires training 5 separate classifier heads for each of the differently stylized images, in addition to a global classifier head which is trained on all the data.
This considerably increases computational memory requirements and training duration.
Despite this, our experimental results with Synthia\textrightarrow Cityscapes demonstrate that our proposed method achieves competitive performance compared to existing approaches like DRPC \cite{Yue_2019_ICCV} and SOMAN \cite{Kundu_2021_ICCV}. It is worth noting that, in our experiments, we utilized a batch size of 1 due to restricted access to computational resources. 
As a future avenue of investigation, exploring the impact of different batch sizes on the GTA5\textrightarrow Cityscapes setting could be valuable to assess their optimal configuration for improved performance and efficiency.

Additionally, it must also be noted that when we use the same batch sizes (SYNTHIA), our performance is considerably better compared to state-of-the-art work \cite{qiaoCVPR20learning,Li_2021_CVPR}.
We suggest that our adversarial stylization method improves training efficiency by generating challenging training examples that are tailored specifically for the segmentation model.
In contrast, both DRPC\cite{Yue_2019_ICCV} and SOMAN\cite{Kundu_2021_ICCV} use stylization methods that do not consider segmentation model performance.}

Furthermore, the pyramidal pooling approach used in DRPC\cite{Yue_2019_ICCV} is a feature-based loss and requires additional computational costs to compare the extracted features across each batch.
In contrast, our approach computes the pixel-wise consistency between the predicted outputs of the original source domain image and the stylized source domain image. 
Since the image output is of lower dimensionality compared to the extracted features, this further reduces the computational costs associated with our method.
In conclusion, our approach demonstrates comparable performance with state-of-the-art work while reducing computation costs.

\subsection{Feature Distribution Visualizations}
{We visualize the feature representations distributions in latent space for a qualitative comparison of the learned representations between our approach and a baseline method.} 
We randomly sampled 10,000 pixels from 40 randomly selected images from the target domain data (\ie Cityscapes) and obtain the extracted feature representations from a baseline model trained only on source domain data (NoAdapt) and a model trained using our method. As seen in \textbf{Figure \ref{figure:tsne plot}}, the feature representations from the different classes for our approach show better separation compared to the feature representations from `NoAdapt'.
\begin{figure*}
\begin{center}
\includegraphics[width=1.0\textwidth,keepaspectratio]{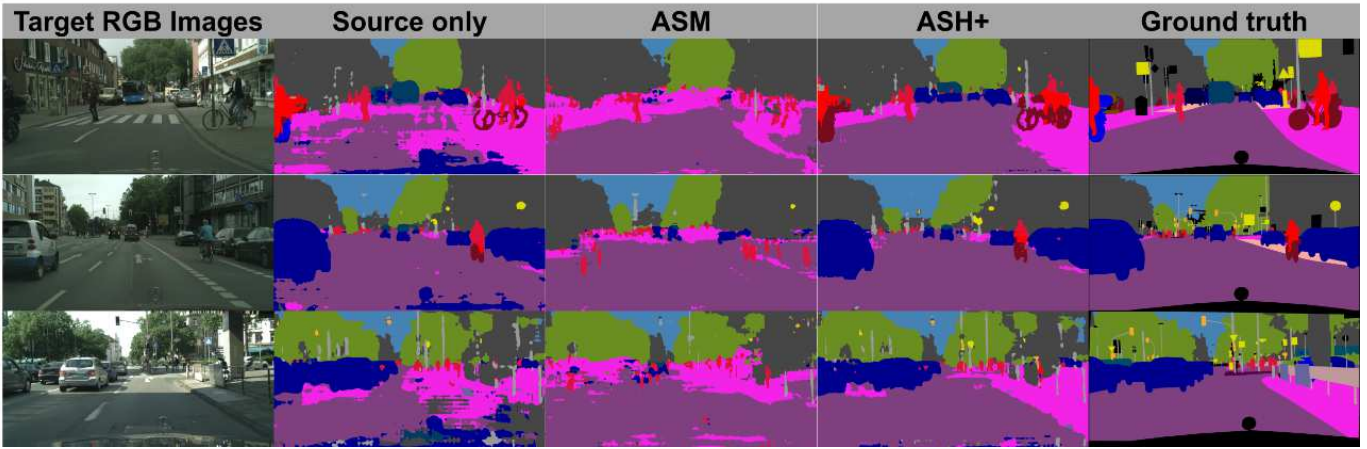}
\end{center}
\caption{Qualitative comparison of segmentation output for Synthia\cite{RosCVPR16}\textrightarrow{} Cityscapes\cite{Cordts2016Cityscapes} For each target domain image, we show the corresponding results for ``Source only", ``ASM" Adversarial Style Mining \cite{DBLP:conf/nips/LuoLGY020}, ``ASH{+}"(our proposed method) and the ground truth labels.}
\label{figure:qualitative_cityscapes }
\end{figure*}

\subsection{Parameter Studies}
\label{subsection:parameterstudies}

We study the effects of varying the hyperparameters $\sigma_{1}$,$\sigma_{2}$, which determines the initial stylization strength.
An optimal choice for $\sigma_{1}$,$\sigma_{2}$ is essential to ensuring that the source domain information is preserved during stylization.
{Recall that $\sigma_{1}$ determines the initial proportion of source domain features in the merged features, while $\sigma_{2}$ determines the initial proportion of stylized source domain features in the merged features.} 

Therefore, we conducted a series of experiments to determine the optimal $\sigma_{1}$,$\sigma_{2}$ values for SYNTHIA as shown in \textbf{Table \ref{tab:synthia highway spring sigma comparison}} and GTA5\textrightarrow Cityscapes as shown in \textbf{Table \ref{tab:gta5 sigma comparison}}.
For the SYNTHIA dataset, we found that the performance is better when a larger proportion of the reconstructed image corresponds to the style features $\boldsymbol{f_{style}}$. 
However, for GTA5\textrightarrow Cityscapes, better task performance was observed when the initial proportions of the style and content features are approximately balanced. 
The difference in optimal $\sigma_{1}$,$\sigma_{2}$ values may be caused by the greater variation in appearance among the SYNTHIA data (e.g.\ Spring \textrightarrow Winter) compared to GTA5\textrightarrow Cityscapes.

The observed performance reduction (approximately 5\%) for different values of $\sigma_{1}$ and $\sigma_{2}$ (see Table \ref{tab:gta5 sigma comparison}) highlights the importance of conducting a thorough search for optimal hyperparameters. 
This search is necessary to ensure that the proposed method performs optimally for each specific dataset.
Furthermore, our experiments have revealed that different datasets may require different values of $\sigma_{1}$ and $\sigma_{2}$. 
For instance, for the SYNTHIA dataset, the optimal values are found to be 0 and 1 (see Table \ref{tab:synthia highway spring sigma comparison}), while for the GTA\textrightarrow Cityscapes dataset, the optimal values are 0.4 and 0.4 (see Table \ref{tab:gta5 sigma comparison}).
Addressing this limitation and achieving an automated, efficient method for hyperparameter search will be a focus of our future research. 
By developing such methods, we aim to improve the adaptability of our approach to different datasets and ensure consistent performance across various domain generalization tasks.

\textbf{Class-wise differences} To further establish the case for class-wise differences in data augmentation, we limited the semantic information from the segmentation output to a single class by providing the $dFT$ layer with the predicted probabilities corresponding to a particular class. 
We then obtained the absolute difference in image intensities between the stylized image and the original source domain image. 
Since the number of pixels predicted may vary widely across classes, we normalized the absolute image intensity difference by the number of predicted pixels for that class. 

As shown in \textbf{Figure \ref{figure:diff_plot}}, the relative intensity difference varies across classes, with larger variations in image intensities for classes such as 'road' compared to those of classes such as 'building'. 
We suggest that 'road' pixels are easier to classify compared to 'building' pixels, causing larger variations during stylization. 

Additionally, $\boldsymbol{\alpha}$ is inversely proportional to the stylization strength since it determines the element-wise proportion of the original source domain features to the perturbed style features (\textbf{Equation \ref{eqn:stylizedsrc}}).
Regions that have high prediction entropy, such as object boundaries, generally have larger $|\boldsymbol{\alpha}|$ compared to regions with lower prediction entropy.
This leads us to suggest that $\boldsymbol{\alpha}$ imposes an upper bound on stylization for regions that are difficult to classify.
\begin{table}[ht]
\setlength{\tabcolsep}{14pt}
\centering
\begin{tabular}{c| c c c c c c c }
\hline
$\sigma_{1}$& 0    & 0.25  & 0.5 & 0.75 \\
$\sigma_{2}$& 1    & 0.75  & 0.5 & 0.25 \\
\hline
Avg. mIoU   &41.39 & 40.54 & 28.37& 20.92 \\
\hline
\end{tabular}
\captionsetup{width=\columnwidth}
\caption{Hyperparameter evaluation with SYNTHIA\cite{RosCVPR16} . The model was trained on Highway-Spring and the performance on the other 10 datasets from New York-like and Old European Town are averaged.}
\label{tab:synthia highway spring sigma comparison}
\end{table}
\begin{table}[h]
\setlength{\tabcolsep}{4.5pt}
\centering
\begin{tabular}{c| c c c c c c c c}
\hline
$\sigma_{1}$& 0.1 &  0.4 & 0.4 &  0.5 & 0.6 &0.7  & 0.75 &  0.25   \\
$\sigma_{2}$&  0.9  & 0.6 & 0.4 & 0.5 & 0.4 &0.3  &  1.5  &  0.5   \\
\hline
mIoU      & 38.62 & 39.04 & \textbf{40.80} & 40.46 & 40.22 &39.41 & 35.70  & 39.61  \\
\hline
\end{tabular}
\captionsetup{width=\columnwidth,justification=raggedright}
\caption{Hyperparameter evaluation for GTA5\cite{Richter_2016_ECCV} \textrightarrow Cityscapes\cite{Cordts2016Cityscapes}} 
\label{tab:gta5 sigma comparison}
\end{table}
\begin{table}[h]
\setlength{\tabcolsep}{7pt}
\centering
\begin{tabular}{c c c c c c c c}
\hline
Baseline & Stylization & Orthogonal &ASH & $\boldsymbol{\alpha}$ & mIoU \\
& & Noise & & & \\
\hline
\checkmark & & & & & 32.38 \\
\checkmark  & \checkmark & & & &  36.24  \\
\checkmark  & \checkmark & \checkmark& & & 37.98  \\ 
\checkmark  & \checkmark &  \checkmark&\checkmark&  & 38.88 \\ 
\checkmark  & \checkmark & \checkmark&\checkmark& \checkmark  &  \textbf{39.80}\\ 
\hline
\end{tabular}
\captionsetup{width=\columnwidth}
\caption{Ablation study for Synthia\cite{RosCVPR16}\textrightarrow{} Cityscapes\cite{Cordts2016Cityscapes}. The baseline approach is the CLAN \cite{Luo_2019_CVPR} method trained on source domain data. Stylization refers to the model trained with additional stylized data, ASH \cite{Tjio_2022_WACV} is our previous work and $\boldsymbol{\alpha}$ refers to the spatial-wise rebalancing factor.}
\label{tab:booktabs_ablation}
\end{table}
We also observe that $|\boldsymbol{\alpha}|$ decreases for classes such as `road' and `building' while remaining relatively consistent for other classes such as 'person'.
This is likely due to the progressively increasing stylization strength for these classes, suggesting that ASH{+} determines stylization strength based on the class information.

\subsection{Ablation Study}
\label{subsection:ablation}
We evaluate the contribution of the different components by conducting an ablation study for Synthia\textrightarrow Cityscapes (\textbf{Table \ref{tab:booktabs_ablation}}).  
Notably, the results improved considerably after introducing ASH{+} with the style-content rebalancing variable $\boldsymbol{\alpha}$. 
Since ASH\cite{Tjio_2022_WACV} previously used an empirically determined global hyperparameter to control the extent of stylization, this further highlights the importance of adaptively modulating the stylization strength.
\section{Conclusion} In this paper, we introduce ASH{+} (Adversarial Style Hallucination+) with the goal of addressing the problem of modulating hallucination strength in domain generalization. 
{By leveraging the semantic information, ASH{+} applies a dual-stage manipulation of hallucination strength.} 
In addition to the learned affine transformation of the style features, we introduce a novel learned {style-content balancing weight} $\boldsymbol{\alpha}$ that is derived from the semantic information.
$\boldsymbol{\alpha}$ balances the proportion of content features and the perturbed style features. 
Experimental results demonstrate the effectiveness of ASH{+}, which yields competitive segmentation performance compared with state-of-the-art domain generalization approaches on two benchmark settings.


%
\bibliographystyle{IEEEtran}
\bibliography{arxiv.bib}

\end{document}